\documentclass[conference]{IEEEtran}
\IEEEoverridecommandlockouts
\usepackage{cite}
\usepackage{amsmath,amssymb,amsfonts}
\usepackage{algorithmic}
\usepackage{graphicx}
\usepackage{textcomp}
\usepackage{dirtytalk}
\usepackage{xcolor}
\usepackage{float}
\usepackage{url}            % simple URL typesetting
\usepackage{hyperref}
\definecolor{mydarkblue}{rgb}{0,0.08,0.45}

\usepackage[caption=false]{subfig}

\hypersetup{colorlinks,linkcolor=red,urlcolor=mydarkblue}

\def\BibTeX{{\rm B\kern-.05em{\sc i\kern-.025em b}\kern-.08em
    T\kern-.1667em\lower.7ex\hbox{E}\kern-.125emX}}
\usepackage{booktabs}
\usepackage{multirow}

\newcommand{\etal}{et al. }

\begin{document}

\urlstyle{tt}

% \title{Undeafening the RL agents, for realsies this time*\\
\title{Agents that Listen: High-Throughput Reinforcement Learning with Multiple Sensory Systems \\
% \title{Agents that Listen: Deep Reinforcement Learning with Multiple Sensory Systems   \\
% \title{Agents that Listen: Deep Reinforcement Learning with Combined Sensor Modalities  \\
% \title{Agents that Hear: Deep Reinforcement Learning from Sound and Vision \\
}

\author{
    \IEEEauthorblockN{Shashank Hegde}
    \IEEEauthorblockA{\textit{University of Southern California}\\Los Angeles, United States \\ khegde@usc.edu}
\and
    \IEEEauthorblockN{Anssi Kanervisto}
    \IEEEauthorblockA{\textit{University of Eastern Finland}\\ Joensuu, Finland \\ anssk@uef.fi}
\and
    \IEEEauthorblockN{Aleksei Petrenko}
    \IEEEauthorblockA{\textit{University of Southern California}\\ Los Angeles, United States \\     petrenko@usc.edu}
\thanks{\textcopyright 2021 IEEE.  Personal use of this material is permitted.  Permission from IEEE must be obtained for all other uses, in any current or future media, including reprinting/republishing this material for advertising or promotional purposes, creating new collective works, for resale or redistribution to servers or lists, or reuse of any copyrighted component of this work in other works.}
}

%\IEEEpubid{\begin{minipage}{\textwidth}\ \\[12pt]
%978-1-6654-3886-5/21/\$31.00 \copyright 2021 IEEE
%\end{minipage}}

\maketitle

\begin{abstract}
    % People have not properly tried adding hearing to the RL agents, despite being a strong input for us apes. Lets fix that.
    % In complex first player games, audio plays an important role in intelligent decision making, yet the application of sound in reinforcement learning is still a relatively new topic of research, with previous research focusing mainly on simple scenarios or navigation. In this work, we set out to introduce a standard environment that can test an AI agent's ability to not only perceive vision, but also hear audio. We build on the popular VizDoom environment, and now provide a state space in sound, that was previously missing. We are now able to provide this in a synchronous fashion, allowing training agents at thousands of frames per second on a commodity hardware. In addition to this, we also study the performance of different encoder networks on this domain and provide baseline results. \todo{link to the webpage.}
    
    Humans and other intelligent animals evolved highly sophisticated perception systems that combine multiple sensory modalities. On the other hand, state-of-the-art artificial agents rely mostly on visual inputs or structured low-dimensional observations provided by instrumented environments. Learning to act based on combined visual and auditory inputs is still a new topic of research that has not been explored beyond simple scenarios.
    To facilitate progress in this area we introduce a new version of VizDoom simulator to create a highly efficient learning environment that provides raw audio observations. We study the performance of different model architectures in a series of tasks that require the agent to recognize sounds and execute instructions given in natural language. Finally, we train our agent to play the full game of Doom and find that it can consistently defeat a traditional vision-based adversary.
    
    We are currently in the process of merging the augmented simulator with the main ViZDoom code repository. Video demonstrations and experiment code can be found at \url{ https://sites.google.com/view/sound-rl}.
\end{abstract}


\section{Introduction}
    % Sound - highly salient, outside of FOV
    % map high-dimensional data to action
    % multiple modalities reinforce each other
    % pro human players use sounds
    % better awareness of 3D surroundings
    % faster learning
    % important source of spatial information for embodied agents
    % real world contains sounds - why don't we simulate them
    Reinforcement learning (RL) algorithms have reached tremendous success in the field of embodied intelligence, including human-level control in Atari games \cite{dqn, badia2020agent57} and in first-person games \cite{vizdoom_competitions, petrenko2020sample}, and super-human control in competitive games \cite{vinyals2019grandmaster, openai2019dota}. These state-of-the-art learning methods allow artificial agents to discover efficient policies that map high-dimensional unstructured observations to actions. While the general framework of deep RL enables learning from arbitrary sources of data, so far the majority of research in embodied AI focused on learning only from visual input (for example, see all the previous citations). We argue that another important sensor modality, sound, is largely overlooked.
    
    Sound represents a highly salient signal rich with information about the environment. Sound cues correspond to discrete events such as contacts and collisions which might be difficult to identify from visual data alone. Stereo sound encodes important spatial information that can reveal objects and events outside of the agent's field of view. Finally, sound could be used to establish a natural communication channel between agents in the form of speech and hearing, which is one of the distinguishing features of higher forms of intelligence.
    
    In computer games, especially in the first-person shooter (FPS) genre, the ability to perceive and understand game sounds is one of the essential skills. This is particularly important in tactical duel scenarios in games like Quake or Doom: in order to gain an advantage skilled players listen to their opponent's actions to understand where they are on the game level and what resources they possess.
    
    Reinforcement learning on combined auditory and visual inputs is complicated by the lack of infrastructure. The existing learning environments either do not support sound, or do not allow high-throughput parallel simulation necessary for large-scale experiments. We attempt to improve the situation by releasing an augmented version of the popular ViZDoom environment \cite{kempka2016vizdoom} where in-game stereo sound is available to the agents. Our implementation is decoupled from dedicated sound hardware typically used for audio rendering, and thus allows faster-than-realtime parallel simulation. We proceed to train agents in our environment in a series of increasingly complex scenarios designed to test various aspects of sound perception.

    \begin{figure}[t]
    \centering
    \centerline{\includegraphics[width=\linewidth]{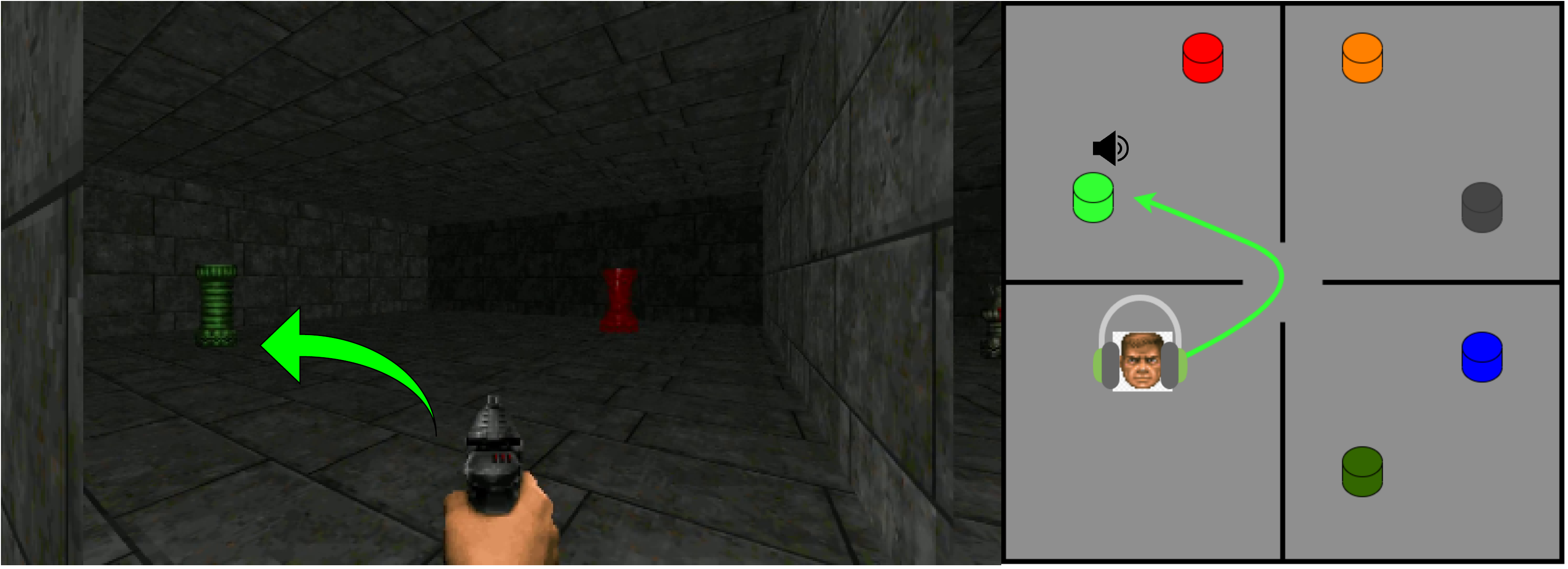}}
    % \vspace{-10pt}
    \caption{Trained agent follows visual and sound cues to reach the target object in a ViZDoom environment.}
    \label{fig:abstract_fig}
        %  \vspace{-5pt}
\end{figure} 
    
    % Anssi: Replaced with paragraph above (using this as the base point)
    % This (ViZDoom) environment has enabled researchers to train considerably formidable AI agents, against bots and self play. These AI agents can defeat many humans but can still fall short of professional Doom players. In an effort to bridge this gap, we explored possible reasons for the AI agent's sub optimal policy. On asking professional players, it was pointed out that sound plays a very important role in a player's ability to make decisions. Therefore, we realized that we need to have access to sound in the state space as well, in order to train better AI bots. This is a difficult task as playing audio (or sound, as it were) is so tightly tied to our "natural passage" of time. In this paper we have developed an environment, AudVizDoom, that lets us do specifically this. We can now synchronously step through a discrete time game while having access to the audio buffer as a part of our state space. In addition to this, we develop scenarios that would be otherwise impossible to solve without the audio state.
    
\section{Related work}
    
    A number of prior projects explored RL with audio observations. Gaina and Stephenson \cite{gaina2019did} augmented General Video Game AI framework to support sound, focusing on 2D sprite-based games. Chen \etal introduced SoundSpaces \cite{chen2020soundspaces}, a version of Habitat environment which focuses on audio-visual navigation in photorealistic scenes. SoundSpaces was further used in \cite{majumder2021move2hear} to investigate the problem of separating sound sources from background noise. Park \etal \cite{park2021veca} introduced a general-purpose simulation platform based on Unity engine with both auditory and visual observations.
    
    While ViZDoom \cite{kempka2016vizdoom} supports in-game stereo sounds, the default audio subsystem is not designed for faster-than-realtime experience collection, and thus can only be used in relatively basic scenarios \cite{woubie2019autonomous}. To our best knowledge, the version of ViZDoom presented in this work is the first simulation platform that enables accelerated embodied simulation with sounds at tens of thousands of actions per second, enabling large-scale training. Our experiments with the Doom duel scenario represent one of the first deployments of an agent with auditory and visual perception in a full first-person computer game.

\section{ViZDoom environment with audio}
    \begin{figure}
        \centering
        \includegraphics[width=\columnwidth]{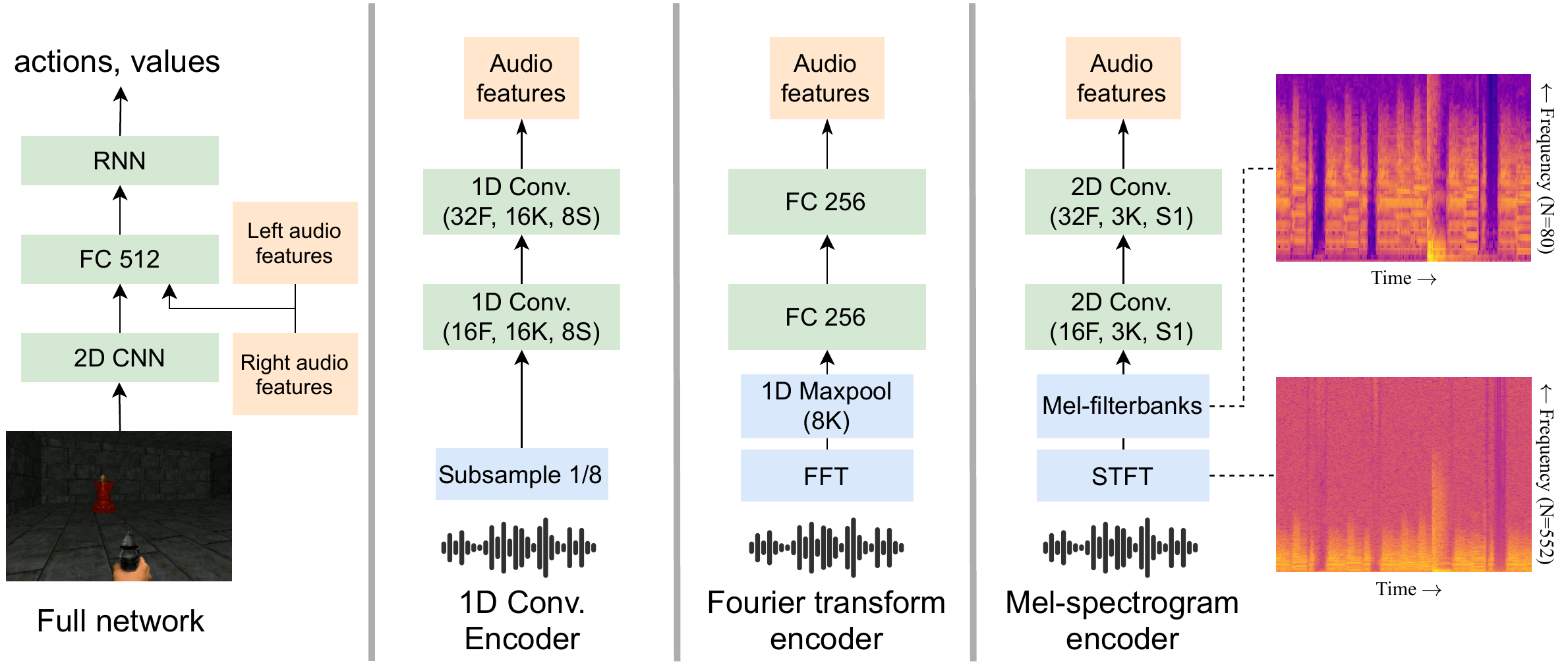}
        \caption{Illustration of the network architecture and audio encoders used (see Appendix of \cite{petrenko2020sample} for complete details). Here K stands for kernel size, F for number of filters, S for stride, FC for fully connected layers and STFT for short-term Fourier transform. All convolutional layers are followed by max-pool layers with kernel size two.}
        \label{fig:audio-encoders}
        %\vspace{-5pt}
    \end{figure}
    
    % We use OpenAL 3D sound
    % We bypass the realtime limitations by using the extension
    % good for parallel sampling
    % frame stacking
    
    We generate the audio observations for the agents through the OpenAL\footnote{\url{https://openal.org/}} sound subsystem supported by ViZDoom. OpenAL implementation offers many modern features, such as 3D sounds, reverberation, Doppler shift, and dampening of the sounds based on the agent's gaze direction with respect to the source.
    
    Normally the sound engine is designed for human perception and plays back the sound samples in real time, prohibiting fast simulation. We circumvent this issue by using OpenAL Soft\footnote{\url{https://github.com/kcat/openal-soft}} with the \textit{ALC\_SOFT\_loopback} extension which completely decouples the in-game sound from the device audio and enables software rendering of sounds on the CPU. This allows us to generate both visual and auditory observations at a maximum rate, enabling the environment simulation in the lock-step fashion typical for a RL setup.
    
    In addition to that, an \textit{ALC\_EXT\_thread\_local\_context} extension allows us to spawn a large number of game instances generating sound samples simultaneously. We leverage that in our experiments by starting hundreds of concurrent processes to achieve high training throughput with an asynchronous RL framework \cite{petrenko2020sample}.
    
    By directly accessing OpenAL sound buffers we expose raw audio observations through ViZDoom API. To give the agents access to all available sound data, we implement configurable audio frame-stacking, independent of the ViZDoom frame-skipping parameters. By default, if the agent chooses its action in the environment once every $N$ simulation steps, we provide the audio observation containing the sounds for the previous $N$ steps. The length of this window can be increased if needed, for example to facilitate training of feed-forward policies.
    
    Another configuration parameter we expose is the audio sampling rate. A larger sampling frequency is analogous to a higher screen resolution, it enables more detailed observations at a cost of increased computation time. In this work we used a fixed sampling rate of 22050 Hz, which provides fast rendering and high sound quality.

    \begin{figure*}[ht]
    \centering
    \subfloat[Music Recognition]{\includegraphics[width=.31\linewidth]{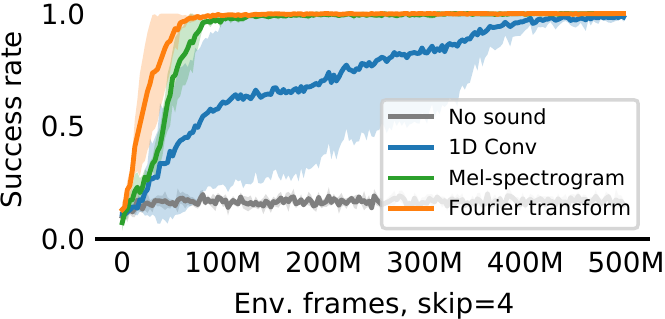} \label{fig:a}}
    \subfloat[Sound Instruction]{\includegraphics[width=.31\linewidth]{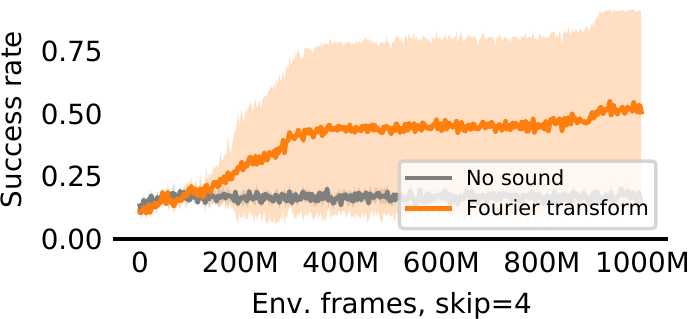} \label{fig:b}}
    \subfloat[Sound Instruction Once]{\includegraphics[width=.31\linewidth]{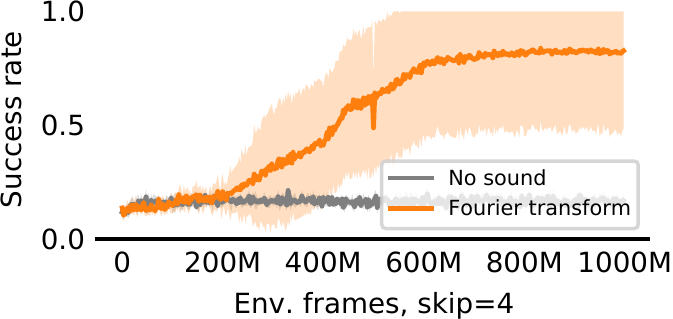}\label{fig:c}}
    \caption{Results on the main testing scenarios. Fig. \ref{fig:a} shows the comparison of different encoders in a sound source finding task. Figs. \ref{fig:b} and \ref{fig:c} show the performance of the FFT (Fourier  transform) encoder on the \textit{Instruction} and \textit{Instruction Once} environments respectively. For each experiment we report mean and standard deviation of five independent training runs.}
    \label{fig:results}
    \end{figure*}

    % AP: original figure caused problems with fonts 
    % \begin{figure*}[t!]
    %     \centering
    
    %     \begin{subfigure}[t]{.32\linewidth}
    %         \includegraphics[width=\linewidth]{plots/doom_sound_instruction.pdf}
    %         \caption{Sound  Instruction}\label{fig:b}
    %     \end{subfigure}
    %     \begin{subfigure}[t]{.32\linewidth}
    %         \includegraphics[width=\linewidth]{plots/doom_once_sound_instruction.pdf}
    %         \caption{Sound  Instruction Once}\label{fig:c}
    %     \end{subfigure}
    %     %\vspace{-3pt}
    %     \caption{Learning curves of different setups, with shaded region representing standard deviation over five repetitions with random seeds. Agent chooses an action every four environment frames. The First curve \ref{fig:a} compares different encoders in a simple sound source finding task.
    %     % Anssi: commented out for space
    %     %The second \ref{fig:b} and third \ref{fig:c} plots show the performance of the FFT (Fourier  transform) encoder on the instruction and instruction once environments respectively.
    %     }
    %     %\vspace{-4pt}
    %     \label{fig:basic_envs}
    % \end{figure*} 

    \begin{figure}[t]
        \centering
        \includegraphics[width=0.4\linewidth]{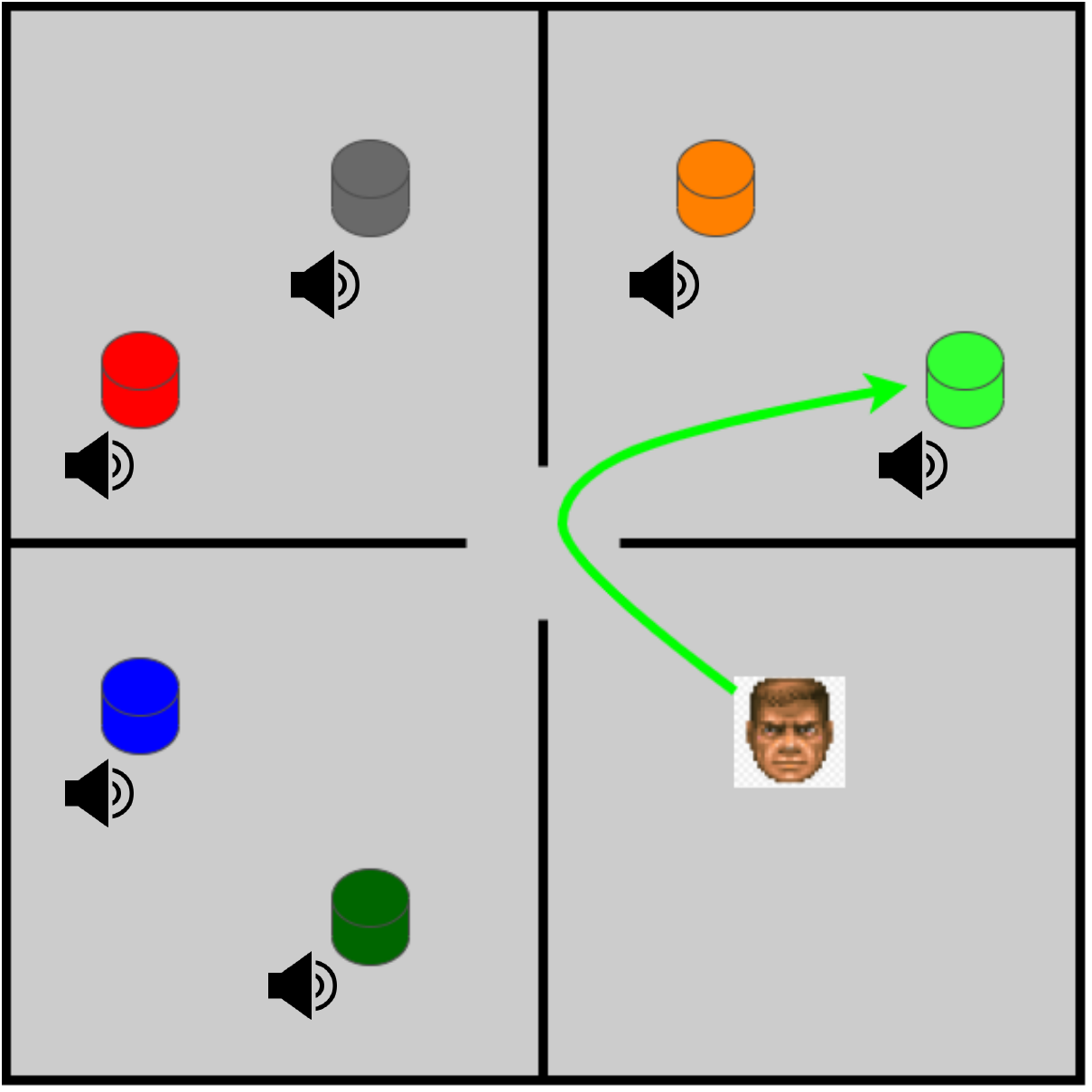}
        \includegraphics[width=0.4\linewidth]{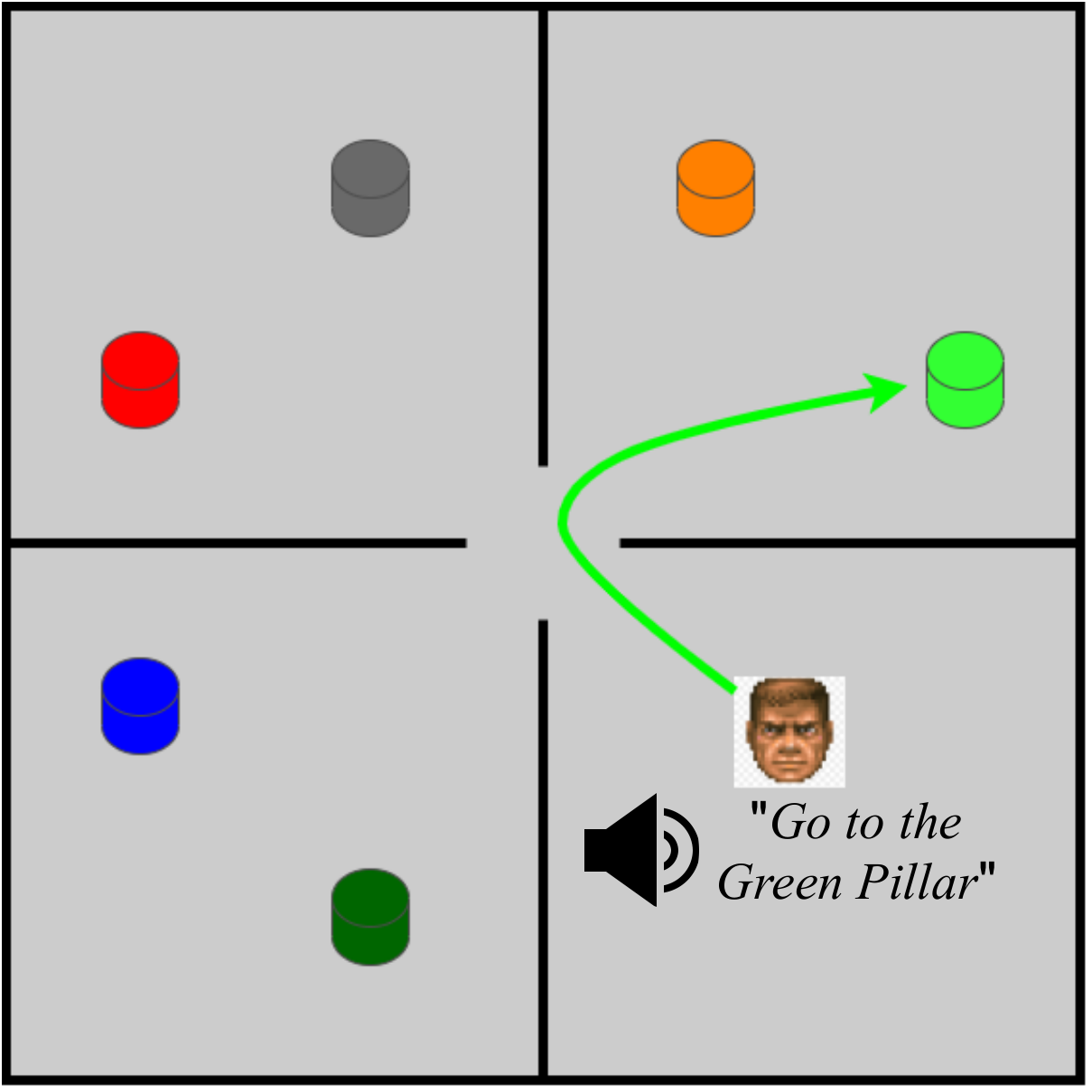}
    
        % \vspace{-10pt}
        \caption{Illustrations of the \textit{Music Recognition} (left) and \textit{Sound Instruction} scenarios (right). Locations of target objects and the player starting position are sampled randomly in each episode.}
        % \vspace{-5pt}
        \label{fig:basic_envs}
    \end{figure} 
    
\section{Audio encoder architectures}
    Our focus is on finding a general approach for processing sound with neural network-based policies. We seek models that are powerful and general enough to solve different, complex tasks, yet compact enough to facilitate fast learning. Using deep learning to process raw image pixels has been successful in RL~\cite{dqn, vizdoom_competitions}, however processing raw audio samples usually takes very large models to do efficiently~\cite{van2016wavenet}, and to this day many state-of-the-art audio systems rely on some form of feature engineering (see Garcia \etal \cite{garcia2020magneto} for an example). These features are applicable to different tasks, with varying levels of performance depending on the task at hand.

    For this reason, we propose three different encoders, which we compare in our experiments. The task of the audio encoder is to generate a compact representation of the raw sound data. This representation is then concatenated with the features from the image processing network. The resulting vector of features is fed to the rest of the network to generate actions and value estimates (see Fig. \ref{fig:audio-encoders}).
    
    The raw audio input is a vector $s \in \mathbb R^{n},~ s_i \in [-1, 1]$ containing $n$ normalized audio samples. ViZDoom runs at a fixed 35 frames per (realtime) second, so for each simulation step this input contains audio corresponding to 29ms of gameplay. With the fixed 22050Hz sampling rate and standard 4-frameskip, our audio observation consists of 2520 samples, or 114ms of audio. We process both left and right audio channels separately and concatenate channel features into a single output vector. Fig.~\ref{fig:audio-encoders} illustrates the high-level structure of the encoders.

    \paragraph{1D Conv.} We downsize the audio by taking every 8\textsuperscript{th} sample and then feed the samples through two 1D convolutional layers. While this removes high-frequency components (anything above $\approx$ 3000Hz), most of the information lies below this frequency threshold. This downsampling allows us to reduce computational complexity. The convolutional encoder can be considered a naive baseline approach
    
    \paragraph{Fourier transform} We transform the audio buffer to frequency domain using fast Fourier transform (FFT) and take the natural logarithm of the magnitudes $s_\text{FFT} = \log \text{FFT}(s) \in \mathbb R^{n/2}$, downsample with a 1D max-pool layer and then feed it through a two-layer, fully connected network. This discards the temporal information inside the 114ms of audio, but enables robust performance and a simple network architecture.
    
    \paragraph{Mel-spectrogram} Motivated by the success of the mel-spectrogram approach in speech processing \cite{garcia2020magneto}, we transform samples into frequency domain spectrogram with short-term Fourier transform (SFTF).
    %$s_\text{spec} = \text{STFT}(s_\text{sound}) \in \mathbb R^{m \times t}$, where $t$ represents number of frames and $m$ is number of frequency bins per window.
    STFT works by sliding a window over the audio samples, computing FFT on that window and then moving the window by a given hop. Depending on the window size and the hop length, the resulting spectrogram can have a large number of feature vectors (depending on the length of the audio) and frequency bins, with most high-frequency components containing only a minimal amount of useful information. Motivated by studies of human audio perception, mel-frequency scale emphasises higher resolution at lower frequencies, usually computed by using triangular overlapping windows \cite{davis1980comparison}. This comes with the additional benefit of reducing the size of the spectrogram. We compute the spectrogram using parameters from \cite{garcia2020magneto}, with a window size of 25ms, 10ms hop size and 80 mel-frequency components. See Fig.~\ref{fig:audio-encoders} for an example of the spectrograms produced with these hyperparameters. The resulting spectrogram is processed by two 2D convolutional layers.
    
\section{Experimental setup}
    
    We train our agents using an asynchronous RL framework Sample Factory. We follow ViZDoom experiments in the original paper \cite{petrenko2020sample} and use the same algorithm, hyperparameters, and model architectures. In particular, we use the asynchronous proximal policy optimization (PPO) algorithm with clip loss \cite{ppo} and V-trace off-policy correction~\cite{impala}. Our model consists of a three-layer, convolutional network to process the RGB image, a chosen audio encoder, a gated recurrent unit layer~\cite{cho2014learning}, and a fully-connected layer to produce action probabilities and value estimates. We ran all our experiments on a single 36-core server with four Nvidia RTX 2080Ti GPUs. 
    %Anssi: Removed this part because it is more discussed later. On single-player testing scenarios with sound the training throughput on this system exceeded $10^5$ environment frames per second (108000 FPS on average).
    
    \subsection{Environment scenarios}
        In order to test the audio encoders and the agent's overall problem-solving abilities we designed three different scenarios based on the map layout depicted in the Figs. \ref{fig:abstract_fig} and \ref{fig:basic_envs}, where six visually distinct pillars are placed in four different rooms. The ordering of pillars and the agent starting position is randomized in each episode. Our fourth and final scenario is a self-play duel in a full game of Doom.

        \paragraph{Music Recognition}
        Each pillar plays a different music track in a loop throughout the episode. One pillar is randomly chosen to play an unique target track. The agent is given a $+1$ reward upon touching the pillar that plays the target track. Touching other pillars terminates the episode with a $0$ reward. The attenuation value of the sound sources is high, therefore the agent has to move close to a pillar to hear the sound. Thus the agent's strategy shall be to move from pillar to pillar and listen, until it finds the pillar playing the target track.
        
        % Here the visual representation of the pillar only acts as a guide for the robots movement, but plays no role in the agent's decision to move towards or away from it.
       
        \paragraph{Sound Instruction} During the episode the agent repeatedly hears a command in spoken English, which instructs it to go to a particular object. The agent is rewarded for touching the correct object and receives zero reward otherwise. Unlike the previous scenario where the decision to move close to an object was purely based on the sound, here the agent has to use both visual and auditory input to complete the task.
        
        \paragraph{Sound Instruction Once} A more complex version of the \textit{Sound Instruction} environment where the instruction is only given once at the beginning of the episode. This scenario tests a combination of multimodal perception and the ability to memorize instructions.

        \paragraph{Duel} Finally, we train our agents in a 1v1 self-play matchup in the full game of Doom, following a setup similar to~\cite{petrenko2020sample} except with full access to in-game sounds. We evaluate the agent against a separately trained agent that is not equipped with the sound encoder, and we hypothesise that the agent with access to sound can outperform the deaf agent.
    
    \subsection{Training settings}
        We test all audio encoders in the \textit{Music Recognition} scenario, where training converges within $5\times10^8$ environment steps. We then choose the best-performing encoder for other experiments, where we train for $10^9$ steps in \textit{Sound Instruction} scenarios and for $2\times10^9$ steps in \textit{Duel} scenario. We fixed the image resolution to 128x72 and set the frameskip to 4 for all environments except Duel which was run with 2-frameskip.

\section{Results}
    After the initial testing on the \textit{Music Recognition} scenario we found that the Fourier transform encoder was the most efficient (Fig. \ref{fig:results}).
    We continued to test the FFT encoder on \textit{Sound Instruction} and \textit{Sound Instruction Once} scenarios. In the majority of the training runs the agent was able to reach optimal performance in each of these scenarios. The high variance in the results suggests that $10^9$ steps of training are still not sufficient for all seeds to converge. We also found that the agent showed slightly better final performance on the supposedly harder task \textit{Sound Instruction Once}. Although it is likely this is just a statistical anomaly given the high variance and low number of independent runs (limited by the computation budget), we leave full explanation of this surprising result for future work.
    
    % Besides, although not statistically s
    % , and we can not draw a definite conclusion on which of these environments is more difficult (no statistical significance, given the high variance). We believe the higher variance in \textit{Sound Instruction} environment is due to sound buffer being constantly filled with audio, which interferes with the hidden state of the GRU layer, while in \textit{Sound Instruction Once} this sound input is only zeros after the initial instruction.

    In the \textit{Music Recognition} scenario, we saw the agent achieve the expected behaviour, where it uses the stereo sound to navigate to the correct pillar. The agent explores the map listening to different music and moves closer to the source when the target music track is recognised. We also saw the agent move towards pillars backwards, showing that visual input is often superfluous in this task. In the \textit{Sound Instruction Once} scenario, we noticed that the agent goes to the center of the map early to await the instructions, which helps minimize the average time to complete the task. While the agent is waiting it keeps turning around memorizing the locations of the objects. Once the instruction starts the agent would quickly turn and approach the target object.
    % The agent even learnt to identify the instruction before it was complete.
    This behaviour shows the agent's ability to combine auditory and visual cues to quickly explore its surroundings and map the sound instruction to the appropriate action. Besides, we noticed the agent's ability to memorize the instructions for the entire episode, courtesy of the recurrent model architecture.
    
    % \begin{table}[t]
    %     \caption{Duel results between an agent that has access to the sound buffer vs an agent that uses just vision, over 100 episodes. ``Sound (dis.)" agent is trained with sound, but sound is disabled during the evaluation.}
    % \centering
    % \setlength{\tabcolsep}{3mm}
    % \small
    % \begin{tabular}{l c c c r}
    
    %     \midrule
    %     Agent 1 & A.1 wins & Draw & A.2 wins & Agent 2 \\
    %     \midrule
        
    %     Sound & 53 & 16 & 31 & No sound \\ 
    %     Sound & 74 & 9 & 17 & Sound (dis.) \\
    %     \bottomrule
    % \end{tabular}
    % \label{tab:duel_results}
    % \end{table}
    
    \begin{table}[t]
        \caption{Results of 1v1 matches between our agent that has access to the sound and a vision-only agent. ``Sound (dis.)" is the main agent with sound inputs disabled during the evaluation.}
    \centering
    \setlength{\tabcolsep}{3mm}
    \small
    \begin{tabular}{l c c r}
    
        \midrule
        \textbf{Match} & \textbf{Wins} & \textbf{Losses} & \textbf{Draws} \\
        \midrule
        
        Sound vs No sound & 53 & 31 & 16 \\ 
        Sound vs Sound (dis.) & 74 & 17 & 9 \\
        \bottomrule
    \end{tabular}
    \label{tab:duel_results}
    \end{table}

     Table \ref{tab:duel_results} shows the benefit of having access to sound information in the \textit{Duel} scenario. Here we trained two sets of agents using population-based training and self-play, with a small population of 4 policies. The main population (\say{Sound}) used the FFT encoder and had access to both auditory and visual observations. Another set of agents (\say{No sound}) used only image observations. After training for $2\times 10^9$ steps we chose the best agents from both populations and ran two series of one hundred 4-minute matches between them. In the first series of games we compared "Sound" and "No sound" versions of the agents. Our main agent won in more games, demonstrating the advantage of the enhanced sensorium.
     In the second series of games we tested our "Sound" agent against a version of itself with its auditory observations replaced with silence (\say{Sound (dis.)}). The agent with disabled hearing played significantly worse, demonstrating the strong reliance of our agent on sound cues.
     
    When analysing the behavior of the main agent in the duel environment we noticed the reduced usage of loud ammunition. We believe this allows the agent to conceal its position from the opponent which facilitates surprise attacks. The agent also uses its spatial sound perception to discover the location of the enemy by listening to the opponent's gunfire.
    
\section{Training throughput}
    To measure the total computation cost added by rendering and processing sound in our experiments we tracked the average training throughput. In single-agent experiments we collected experience using 72 parallel workers, each worker sampling 8 environments sequentially for a total of $72 \times 8 = 576$ parallel environments per experiment. We ran 4 such experiments at a time on a 36-core machine with 4 GPUs to maximize the hardware utilization. We observed training throughput of $1.5 \times 10^5$ game frames per second per experiment with disabled sounds and $1.2 \times 10^5$ when sound is enabled. 
    
    We did not notice a significant difference in performance in Duel scenario. Here we trained a population of 4 policies at a combined framerate of $6.7 \times 10^4$ both with and without the sound. The training performance in multi-agent VizDoom envs is bottlenecked by slow network-based communication between game instances in the multi-agent setup, and thus addition of sound rendering workload does not have a significant effect.
    
\section{Conclusions and future work}
    In this work we introduced an immersive environment based on ViZDoom that provides access to both auditory and visual observations while maintaining high simulation throughput. We introduced new scenarios that test the agent's ability to hear and identify sounds, as well as combine sound with visual cues. Our results indicate that transforming the audio samples into frequency domain with FFT is sufficient for fast and effective RL training when combined with a recurrent neural architecture. This is evident from the results of our experiments with sound separation and instruction execution, as well as results on a full game where the agents with augmented sensorium prevail. We hope that access to the efficient environment that simulates auditory experience will enable large-scale experiments and can facilitate further research in this area.
    
    Being a preliminary work, there are still a myriad of open questions and limitations to address. We only used one RL algorithm in our experiments and only three different audio encoders, without excessive hyperparameter and/or architecture tuning. The scenarios used in our experiments could also be extended: we used a limited bank of sounds in the experiments, and to assess how well the agent learned to \say{understand sound} instead of overfitting to specific cues, we need a larger bank of sounds to pick from. This can be done by adding more natural sounds and by augmenting the existing ones with random noise and other transformations to prevent the neural network from memorizing the exact samples.
    
    % We also only assessed the agent's ability to detect one sound from others and did not evaluate its ability to localize the sound source. This may be present in the duel experiment conducted in this work, however this should be studied in depth as well. Currently we only evaluated the performance of the self-play agents, but we should also study why the hearing-agent was able to outperform the vision-only agent.
    
    While it is evident from our experiments that the agents benefit from the addition of auditory observations, it is not clear how exactly the agents utilize the sound cues. For our agents trained in the Duel scenario the behavior of the hearing agents can be studied in-depth, i.e. to find out in what ways the agent utilizes the sound and how well it can localize its opponent in 3D. We leave this interesting research direction for future work.

    Finally, while this work used a recurrent neural architecture for temporal modelling (\say{memory} for the agent), it is unclear whether agents can understand long audio sequences. One should assess this with, for example, longer-lasting audio cues or longer, more dynamic commands (e.g. not only \say{go to X}, but also \say{do not go to X}, etc.) Preferably, these experiments should be combined with other cognitive tasks like in DMLab30 \cite{impala} to evaluate the agent's ability to really understand sound information. This could be compared to a pipelined baseline approach, where audio is preprocessed through a speech recognition system to assess the agent's ability to learn language understanding with pure end-to-end RL.

\section*{Acknowledgments}
    We thank the reviewers of this paper for very insightful comments and ideas for the future work, which we included in the previous section.

\bibliographystyle{ieeetr}
\bibliography{main}

\end{document}